# A Multi-Size Neural Network with Attention Mechanism for Answer Selection

**Jie Huang**

National University of Defense Technology
Correspondence: haungjie_gh@126.com

**Abstract:** Semantic matching is of central significance to the answer selection task which aims to select correct answers for a given question from a candidate answer pool. A useful method is to employ neural networks with attention to generate sentences representations in a way that information from pair sentences can mutually influence the computation of representations. In this work, an effective architecture -- multi-size neural network with attention mechanism (AM-MSNN) – is introduced into the answer selection task. This architecture captures more levels of language granularities in parallel, because of the various sizes of filters comparing with single-layer CNN and multi-layer CNNs. Meanwhile it extends the sentence representations by attention mechanism, thus containing more information for different types of questions. The empirical study on three various benchmark tasks of answer selection demonstrates the efficacy of the proposed model in all the benchmarks and its superiority over competitors. The experimental results show that (1) multi-size neural network (MSNN) is a more useful method to capture abstract features on different levels of granularities than single/multi-layer CNNs; (2) the attention mechanism (AM) is a better strategy to derive more informative representations; (3) AM-MSNN is a better architecture for the answer selection task for the moment.

**1. Introduction**

Question answering (QA) system aims to meet users' requirements by returning a list of answers according to users' question. With the development of intelligent robot, intelligent customer system, smart home and so on, QA has been paid more and more attention to the natural language process (NLP) and machine learning (ML) communities (Ferrucci [1]; Etzioni [2]). Answer selection (AS) is a crucial subtask of QA system and lots of other tasks such as machine comprehension. In recent years, the problem has been receiving an increasingly amount of attention [3-16] Given a question q and a candidate answers pool $P = \{a_1, a_2, ..., a_p\}$, the goal is to search for the correct answer $a_k$, where $1 \leq k \leq p$. Answers and questions are of flexible lengths, and a question can correspond to multiple ground-truth answers. If the selected answer $a_k$ is inside the ground truth answers set, the question q is considered to be answered correctly, otherwise not. The answers in candidate pools are sampled randomly from the whole answers. In testing, the candidate answers for questions are not observed in the training phase.

The major challenge of this task is that researchers need to match questions and answers at meaning level, not just at lexical level. Because the correct answers may only relate with questions semantically, without sharing the same lexical units. Therefore, they propose various approaches to solve this problem.

Most prior work on this task deals with it by relying fully on manually designed, task-specific linguistic features, called traditional methods in the present paper. Traditional methods are based on lexical features such as parsing tree edit distance. But they require feature engineering, linguistic tools or external resources which may be quite difficult to obtain in practice and are error-inclined. Recently, deep learning methods such as CNNs and RNNs are proposed and applied to a host of fields. They simulate the human brain's work mechanism by employing deep neural networks with activation function, loss function and optimizer to learn features automatically without lots of domain knowledge. With these approaches, the question and answer sentences can be represented into a common hidden vector space by capturing sentence features and max/average-pooling operations. For example, In [9, 11], the authors proposed single-layer CNN (SingleCNN) and two-layer CNN (MultiCNNs) to generate the sentence representations, but they did not find improvement in the latter structure. This can be explained by different granularities the filters extract from sentences.

However, these models represent questions and answers separately, which may ignore the information subject to the question when representing the answer, and vice-verse. This absence of the interactive influence of the two sentences in the context contradicts what humans do when we estimate whether one sentence relates to another. Humans usually focus on certain parts of one sentence by using their relation such as identities, synonymies, antonyms and so on. In other words, human beings compare two sentences together instead of separately, by using the contextual information between two sentences. For example, there is a triplet$< q, a+, a->$:

q **: How much is J. K. Rowling worth?**

a $+$ **: The 2008 Sunday Times Rich List estimated Rowling 's fortune at £560 million ($798 million), ranking her as the twelfth richest woman in the United Kingdom.**

a $-$ **: The Potter books have gained worldwide attention, won multiple awards, and sold more than 400 million copies.**

Here q, a$+$ and a$-$ refer to the question, the ground truth answer and the negative answer, respectively. Correctly answering q requires putting attention on 'worth'. The answer a$+$ contains a corresponding unit ('fortune' and 'richest') while a$-$ does not. Therefore, researchers try to propose different attention mechanisms into answer selection task to emphasize certain parts of sentences, such as [7-10, 17, 18].

In order to mitigate the impact of different granularities, the multi-size neural network (MSNN) is introduced into answer selection task in present paper, because it contains different sizes of filters that can be used in parallel. As a result, it can capture features on multiple levels of granularities, not only at the word level, but also at phrase level, even at tri-gram level or more. By virtue of this outstanding property, the method in this paper performs much better comparing with single-layer CNN, multi-layer CNNs and biLSTM, which is a typical RNN frequently used in exploiting long-range dependencies problems.

Meanwhile, a new attention mechanism (AM) is proposed in the present paper. It derives more informative representations by Hadamard product (element-wise product) and concatenation operations, thus improving its performance.

Finally, MSNN and AM are combined (thus AM-MSNN) to make full use of the strongpoints of MSNN and AM. Experimental results show that this combination achieves superior performance, comparing with state-of-the-art single-layer CNN, multi-layers CNN and biLSTM.

Experiments in the paper are conducted with three publicly available benchmark datasets, which vary in data scale, complexity and length ratios between question and answers: InsuranceQA, TREC-QA and WikiQA. The empirical study demonstrates that (1) a useful way to capture abstract features on different levels of granularities, which benefits sentence modeling, is to utilize MSNN. (2) AM is a better strategy to generate more informative representations for two reasons: (a) it employs Hadamard production and column-wise max pooling operations to calculate the weighted representations, which has the potential to search out the biggest concern for q, a$+$ and a$-$; and (b) it uses concatenation operation over the weighted representations and the outputs of networks to take more information into account; (3) AM-MSNN is a good choice for answer selection task.

The structure of this paper is shown as follows: Section 2 discusses related work. Section 3 introduces multi-size neural network (MSNN), attention mechanism (AM) and the combination architecture (AM-MSNN). Sections 4 details experiments and discussion. Section 5 concludes this work.

## 2. Related Work

This section states the approaches for answer selection, including traditional methods, deep learning and attention-based neural networks. In the end, the development of PyTorch will be briefly introduced which is utilized to implement neural networks.

*2.1. Traditional Methods on NLP*

Due to the variety of word choices and inherent ambiguities in natural languages, bag-of-word approaches with simple surface-form word matching tended to produce brittle results with poor prediction accuracy [19]. As a result, researchers paid more attention to exploiting syntactic and semantic structure. For example, methods based on deeper semantic analysis were proposed in [20]. Some works tried to fulfill the matching using minimal edit sequences between dependency parse trees [21]. In [22], the answer selection problem was transformed to a syntactical matching between the question/answer dependency parse trees. Instead of focusing on the high-level semantic representation, Yih, *et al.* [3] used semantic features from WordNet to pair semantically related words based on word semantic relations. Lai and Hockenmaier [23] explored more fine-grained word overlap and alignment between two sentences using negation, hypernym, hyponym, synonym and antonym relations. Yao, et al. [24] extended word-to-word alignment to phrase-to-phrase alignment by a semi-Markov CRF (conditional random fields).

Although these methods are effective, there are still some drawbacks. For example, they require high-quality data and various external resources which may be quite difficult to obtain. Linguistic tools, such as parse trees and dependency trees, increases systematic complexity, resulting in consuming more computational resources. There are inescapable errors of many NLP tools such as dependency parsing because of a large amount of human work.

*2.2. Deep Learning Approaches on NLP*

Recently, Deep Learning (DL) approaches had been widely used on various natural language processing tasks, such as semantic analysis [25] and machine translation [26], often taking on superior performance compared to traditional methods. CNNs (Convolutional Neural Networks, [27]) based network for sentence matching [28], sentiment prediction and question classification [29] and sentiment analysis [30] were proposed continuously. A recursive network was employed for paraphrase detection and parsing in [31].

For AS, deep learning methods were utilized to learn the distributed representation of question-answer pair directly in a same hidden space and achieved state-of-the-art performance compared to traditional non-DL methods. A Deep Belief Nets (DBN) was proposed to learn the distributed representation [32]. Then, a bigram CNN [5] to model question and answer sentences was employed, which was extended and tested on WikiQA dataset in [33]. Different setups of a bi-CNN were tested on an insurance domain QA dataset in [11]. Bidirectional LSTM (biLSTM) was explored on the same dataset in [10]. One common point is that the representations of questions and answers are generated separately, and score a QA pair using a similarity metric to select the highest score as the ground truth answers. In [34], joint feature vectors connecting questions and answers were learned from LSTM model (long short-term memory,[35]) and tested on TREC-QA dataset. Hu, *et al.* [28] presented 2-dimensional convolutional models to represent the hierarchical structures of sentences which captures rich matching patterns.

2.2.1. Single-layer CNN and Multi-layer CNN

Convolutional Neural Networks (CNN) are different from fully connect networks with three important ideas to improve its performance: sparse interaction, parameter sharing and equivariant representation. Sparse interaction contrasts with traditional fully connect neural networks where next layer neurons connect all the ones in above layer, resulting in every output neuron is interactive with every input neuron. In CNN, there is a filter which is usually much smaller than the input size to scan all the input so that the output is only influenced by a narrow window of the input in the scope of the filters each time. Parameter sharing refers to the filter parameters similar to the weight matrix elements of fully connect neural networks and are reused in the convolution operations, while the ones in fully connect neural network only used once to calculate the output. Equivariant representation refers to the idea of k-MaxPooling which is added at the end of CNN to make sure that it can gets the best features. In the present paper, k is equal to 1 for models MSCNN, and AM is employed to take the place of MaxPooling operation.

Multi-layer CNN [9, 11, 36] is proposed to extract high level features, because too fewer layers of convolutions cannot get sufficient abstracted information especially in image recognition domain. In this paper, a two-layer CNN is employed for sentence modeling because researchers in [9] had discussed too many layers' convolutions have no contributions for answer selection.

2.2.2. LSTM and biLSTM

Recurrent Neural Networks (RNNs) have been widely exploited to deal with variable-length sequence inputs [10, 34]. However, with the discovery of the gradient vanishing problem, it is difficult for conventional RNN to exploit long-range dependencies. LSTM is one of the earliest variations of RNN and still a popular method to mitigate this problem. In this paper, LSTM implementation is adopted, similar to the one in [37] with minor modification. In the architecture, LSTM uses input gates and output gates to control the flow of information through the cell and uses forget gate to reset the cell's own state. The long-distance information is stored in a recurrent hidden vector which is dependent on the immediate previous hidden vector.

Single direction LSTM suffers a weak point of utilizing only previous context with no contextual information from the future context. Bidirectional LSTM (biLSTM) utilizes both the previous and future context by processing the sentences from two directions with two separate hidden layers, and generates two independent output vectors. One processes the input sequence in the forward direction, while the other processes the input in the backward direction. The both output vectors from two directions are concatenated as the output at each time step, i.e. $h_t = \overrightarrow{h_t} \parallel \overleftarrow{h_t}$.

*2.3. Attention-Based NN in NLP*

One obvious drawback of the approaches above is that sentences are modeled separately, which ignores semantic correlations between questions and answers. This does not conform to what humans think when they estimate whether one sentence relates to another. Specifically, human beings model sentences by integrating each other instead of independent items and focusing on specific parts related to questions automatically by so-called attention mechanism.

Weston, et al. [38] employed an external memory to store the knowledge. The memory with respect to the attention, called Memory Networks, is used for following inference. Then, lots of variants were introduced by many researchers such as [7-10, 17, 18]. In [10], an attentive reader based on bidirectional long short-term memory was developed, which employed the question representations to guide the answer's generation by emphasizing certain part of the answer. Inspired by this idea and in order to mitigate attention bias problem, a series of inner attention mechanisms based gated recurrent unit (GRU) was proposed. They introduced attention information before RNN hidden representation by paying attention to the original input words, previous hidden state and the GRU inner activation.

The mentioned methods are one-way attention mechanisms, which guide answer representation by question representation. Rocktäschel, *et al.* [18] proposed a two-way attention method based RNNs which made the generation of questions and answers representations interact each other. Yin, *et al.* [9] presented a two-way attention mechanism that is tailored to CNNs [11]. dos Santos, *et al.* [7] also proposed a two-way attention mechanism which used similarity metric learning instead of Euclidean distance in [9] to compute the interdependence between the two input texts.

Recently, Wang, *et al.* [6] proposed a hybrid model which used the four outputs of max-pooling and min-pooling operations on the word embedding of the original answer and the augmented question, the concatenation of the original question and the answer, as the final representations. And then it computes the cosine score between the two outputs of max-pooling and min-pooling operations, respectively. In [17], a structured self-attentive architecture was proposed, which used two learning matrix and self-representation to generate weighted representations and achieved state-of-the-art performance.

*2.4. PyTorch*

In March 2017, based on Torch, a machine learning and scientific computing tool, Facebook released PyTorch [39], another new machine learning toolkit for the Python language. Once it was released, the open source toolkit has received widespread attention and discussion. After several months of development, PyTorch has now become one of the most important R & D tools for practitioners. Because of the auto grad mechanism, PyTorch constructs neural networks with strong GPU acceleration in a dynamic way. Comparing to the static way in TensorFolw, it is faster and more flexible. Meanwhile, PyTorch makes codes simple and succinct.

**3. Methodology: Multi-Size Neural Network with Attention Mechanism (AM-MSNN)**

This part formally describes the proposed framework: MSNN, AM and AM-MSNN. MSNN aims to mitigate the impact of different granularities. AM derives informative representations for sentences. AM-MSNN benefits the strongpoints of both MSNN and AM. And then the similarity metric to measure the distance of question answer pairs is described.

*3.1. The Whole Architecture of Multi-Size Neural Network with Attention Mechanism*

**Figure 1** presents the joint illustration of the whole architecture of AM-MSNN for answer selection.

**Inputs:** Given the input pair $(q, a)$, where the question $q$ contains M tokens and the candidate answer $a$ contains N tokens (M = N in our actual architecture for sentences are padded or truncated to a fixed length for calculation speed). Let $x_i$ be the d-dimensional word vector corresponding to the i-th word in the sentence. A sentence $S = (w_1, w_2, ..., w_s)$ in the input pair, which has s tokens, is represented as

$$x_{1:s} = x_1 \oplus x_2 \oplus ... \oplus x_s. \qquad (1)$$

Here $\oplus$ is the concatenation operator.

**Word embedding:** Let $x_{i:i+j}$ refer to the concatenation of words $x_i, x_{i+1}... x_{i+j}$. Word embedding operation transforms each input word $w_i$ into a fixed-size real-valued word embedding $x_i \in \mathbb{R}^d$. Word embedding are encoded by column vectors in an embedding matrix $W \in \mathbb{R}^{d \times |V|}$, where V is a fixed-sized vocabulary and d is the dimension of the word embedding.

**Multi-Size Neural Network (MSCNN):** Next these sentences are processed by multi-size neural network. The convolution operation involves a filter $c \in \mathbb{R}^{kd}$, which is applied to a window of k words to produce a new feature. Then a feature $f_i$ is generated from a window of words $x_{i:i+k-1}$ by

$$f_i = f(w \cdot x_{i:i+k-1} + b), \qquad (2)$$

where $b \in \mathbb{R}$ is a bias term and $w \in \mathbb{R}$ is the weight, which will be updated in the training process. Here $f()$ is a non-linear function such as sigmoid, tanh and Relu. This filter will be applied to all possible window of words in the sentence $\{x_{1:k}, x_{2:k+1}, ..., x_{s-k+1:s}\}$ to generate a feature map

$$f = [f_1, f_2, ..., f_{s-k+1}], \qquad (3)$$

What has been described above is the process that one feature is extracted by one filter. MSCNN uses multiple filters with varying filter sizes to obtain different features $Q_f \in \mathbb{R}^{c \times M}$ and $A_f \in \mathbb{R}^{c \times N}$, where c is the number of convolutional filters.

**Attention Mechanism (AM):** Then, AM outputs final distributed representations $r_q$ and $r_a$.

**Cosine Similarity:** Finally, cosine similarity is utilized to measure their distance and then rank the answers.

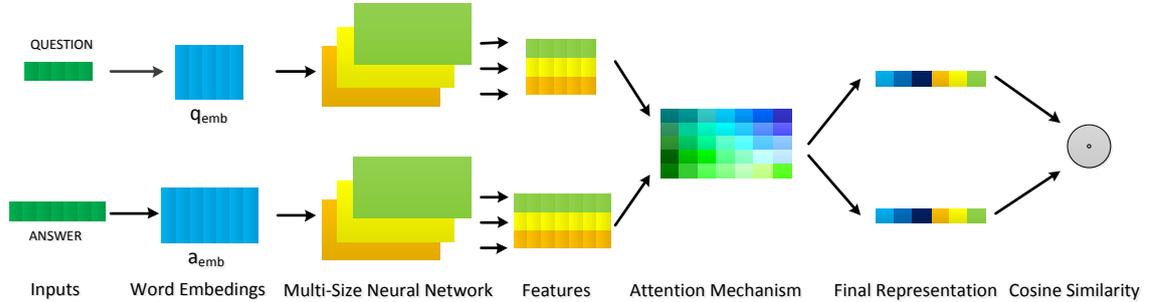

**Figure 1.** The architecture of multi-size neural network with attention mechanism (AM-MSNN).

*3.2. Multi-Size Neural Network (MSNN)*

In order to take language granularity into account, multi-size neural network (MSNN) is proposed to adapt to the answer selection task. For MSNN, convolutional filters with varied sizes are employed to capture different abstraction features. As is showed in Figure 2, MSNN is composed of three groups of convolution operations with filter sizes 1, 3 and 5. Abstract feature maps are produced by concatenating all convolutions outputs. Firstly, embedded sentences are processed by 1*100 or 1*300 convolution filters (matched to word embedding size). It not only reduces computational complexity, but also captures abstraction features on the granularity of one-word level, because the filters scan the whole sentence word by word without any overlap. Since directly applying more convolutional filters and concatenating them are computationally expensive, 1*100 convolutional filters at the bottom of the network is used, which reduces the input dimensionality and hence decreases the computation cost greatly. Secondly, filters with size of 3*3 (padding=1) and 5*5 (padding=2) are employed to capture the features for tri-gram level and 5-gram level of granularity. By this carefully crafted designing, MSNN increases the depth and width of the network to capture more features for various levels of language granularity, while keeping the computational budget constant. In the end, after pooling operations, all abstract features are concatenated together to get features for sentences.

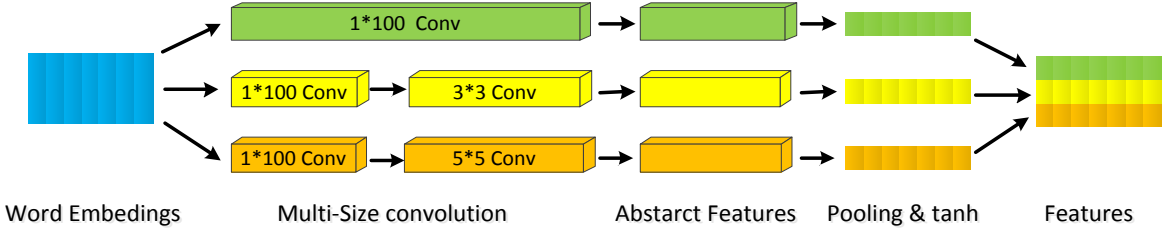

**Figure 2.** Multi-Size Neural Network (MSNN).

The whole convolution operation is described as follows. Given a set of filter with various sizes $C = \{c_1, c_2, \ldots, c_t\}$, where $c_i$ is corresponding to the i-th filter in the filter set. According to formula (2) and (3), outputs of the i-th filter can be obtained,

$$f_i^{c_i} = f(w^{c_i} \cdot x_{i:i+k-1} + b^{c_i}), \qquad (4)$$

$$f^{c_i} = [f_1^{c_i}, f_2^{c_i}, \ldots, f_{s-k+1}^{c_i}], \qquad (5)$$

Next, in order to capture the most important feature for each feature map, a max pooling operation is applied over the feature maps and the maximum value is the feature corresponding to the filter,

$$\widehat{f^{c_i}} = \max\{f^{c_i}\}. \qquad (6)$$

The features of the input sentence are represented as,

$$\hat{f} = \widehat{f^{c_1}} \oplus \widehat{f^{c_2}} \oplus \ldots \oplus \widehat{f^{c_t}}. \qquad (7)$$

Then the sentence representation $Q_f = \hat{f}$ or $A_f \in \hat{f}$ can be obtained, which depends on which sentence in the input pair $(q, a)$ is passed to MSCNN.

MSNN captures more kinds of levels of granularities in a way different from multi-layer CNN. In multi-layer CNN, the level of granularity depends on the width of filter and the number of layers, such as trigram corresponding to the filter width w = 3 in first layer, 5-gram corresponding to the filter width w = 3 in second layer (5 = 1 + 2 * (3 - 1)). Comparing to the single-layer CNN, the process of features capturing on different levels of granularities is the sub-process for first layer in multi-layers CNN.

MSNN is based on the fact that the more different abstraction features captured; the more information can be obtained in final sentence representations. And MSNN gives the network more samples to learn.

*3.3. Attention Mechanism (AM)*

Here attention mechanism is detailed, which models sentences by integrating the information from questions and answers. **Figure 3** shows the basic attention architecture.

After MSNN, the features $Q_f$ and $A_f$ are obtained which are captured by convolution operations. Each column is the representation of a unit. Then, the attention matrix $T \in \mathbb{R}^{M \times N}$ is computed as follows,

$$T = \tanh(Q_f^T U A_f), \qquad (8)$$

where $U \in \mathbb{R}^{c \times c}$ is a matrix of parameters to be learned in training. The attention matrix $T$ contains the scores of a soft alignment between each units in q and a.

Then column-wise and row-wise max-pooling operations are applied over $T$ to derive weight vectors $g^q \in \mathbb{R}^M$ and $g^a \in \mathbb{R}^N$, respectively. Formally, the j-th elements of the vectors $g^q$ and $g^a$ are computed as follows,

$$[g^q]_j \in \max[T_{j,m}], 1 < m < M, \qquad (9)$$

$$[g^a]_j \in \max[T_{l,j}], 1 < l < N. \qquad (10)$$

Next, the softmax function is applied on vectors $g^q$ and $g^a$ to create attention vectors $\sigma^q$ and $\sigma^a$, respectively,

$$[\sigma^q]_j = \frac{e^{[g^q]_j}}{\sum_{1<i<M} e^{[g^q]_i}}, \qquad (11)$$

$$[\sigma^a]_j = \frac{e^{[g^a]_j}}{\sum_{1<i<N} e^{[g^a]_i}}. \qquad (12)$$

Each element j in attention vectors $\sigma^q$ corresponds to the j-th unit in question q and it represents the importance score of this unit with considering the information from answer a. Likewise, each element j in attention vectors $\sigma^a$ corresponds to the j-th unit in answer a and it represents the importance score of this unit with considering the information from question q.

Then, the features $Q_f$ and $A_f$ for each unit are reweighted by attention vectors $\sigma^q$ and $\sigma^a$. In this way, the restriction relationship between q and a is shipped, and the biggest concern for question q with regard to answer a is searched out. Thus, Hadamard production is conducted to update the outputs of convolution $Q_f$ and $A_f$ with the attention vectors $\sigma^q$ and $\sigma^a$, respectively,

$$Q_{f\_weighted} = Q_f * \sigma^q, \qquad (13)$$

$$A_{f\_weighted} = A_f * \sigma^a, \qquad (14)$$

where * is the Hadamard production operator. And then the updated outputs of convolution $Q_{f\_weighted} \in \mathbb{R}^{c \times M}$ and $A_{f\_weighted} \in \mathbb{R}^{c \times N}$ are obtained.

Succeeding the operation, column-wise max pooling operations are applied on $Q_{f\_weighted}$ and $A_{f\_weighted}$ to search out the most concerned units in question q (or answer a) with considering the information from answer a (or question q). And then the weighted representations $r_{weight\_q}$ and $r_{weight\_a}$ for question q and answer a are derived, respectively.

In order to utilize the information contained in outputs of convolution, column-wise max pooling operations are applied over the outputs of convolution and the most representative feature $r_{q\_without\_att}$ and $r_{a\_without\_att}$ is obtained, corresponding to each particular filter. Then, the weighted representations $r_{weight\_q}$ and $r_{weight\_a}$ are concatenated with $r_{q\_without\_att}$ and $r_{a\_without\_att}$, respectively,

$$r_q = r_{q\_without\_att} \oplus r_{weight_q}, \qquad (15)$$

$$r_a = r_{a\_without\_att} \oplus r_{weight\_a}. \qquad (16)$$

Finally, representations $r_q$ and $r_a$ are obtained, which are considered the final representations for input pair. And this attention mechanism makes sure that all information captured by neural networks and attention has been taken into account. Meanwhile, this scheme naturally deals with variable sentence lengths.

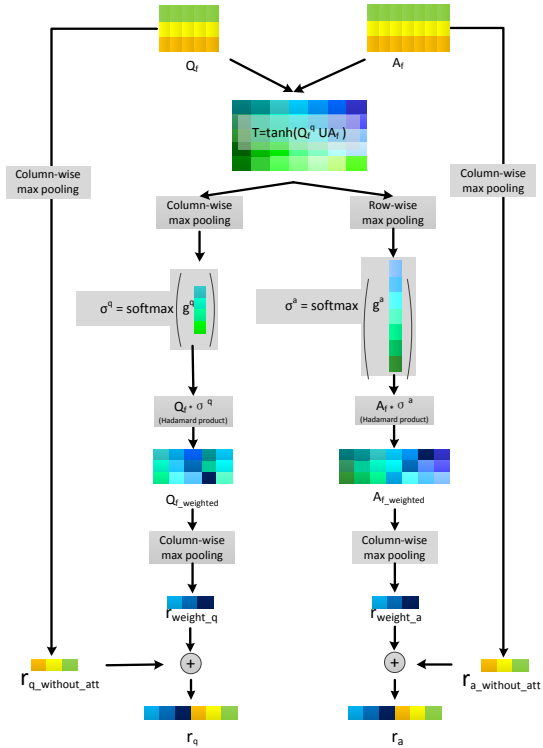

**Figure 3.** The attention mechanism architecture (AM).

*3.4. Cosine Similarity and Training Procedure*

The architecture utilizes cosine similarity to measure their distance in the end. During training, for each training question q, there is a positive answer a +. A training instance is constructed by pairing this a + with a negative answer a − in candidate answer pools. The whole framework generates the representations for the question and the two candidates: $r_q$, $r_{a+}$ and $r_{a-}$. The cosine similarities $\cos(r_q, r_{a+})$ and $\cos(r_q, r_{a-})$ are calculated as follows,

$$\cos(q, a) = \frac{r_q \cdot r_a}{||r_q|| \cdot ||r_a||}, \quad (17)$$

where $||\cdot||$ is the length of the vector.

In order to minimize the pairwise ranking loss function, according to [11, 28, 38], a max-margin hinge loss function is adopted as training objective,

$$L = \max\{0, m - \cos(q, a+) + \cos(q, a-)\}, \quad (18)$$

where $m$ is constant margin. For testing, all $\cos(q, a_{candidate})$ between the question and every candidate answer are calculated. The candidate answer with largest cosine similarity is selected according to cosine similarity.

## 4. Experiments

### 4.1. Datasets

In this paper, a comprehensive set of experiments are presented over three QA datasets: InsuranceQA, WikiQA and TrecQA. These datasets vary in data scale, complexity and length ratios.

InsuranceQA [11] is a recently released large-scale non-factoid QA dataset from the insurance domain. It consists of a training set, a development (dev) set, and two test sets. For each question in dev or test sets, a set of 100/500/1000 candidate answers is provided in latest version, including the ground-truth answers and negative answers randomly sampled from the whole answer space. The set of 500 candidate answers in the old version is chosen for experiments in this paper because the baseline experiments are at this level [7, 8].

WikiQA [33] is an open-domain question answering dataset which contains 3,047 questions originally sampled from Bing query logs. The corresponding dataset consists of 20,360 questions candidate pairs (a question corresponds to multi answers) in train, 1,130 pairs in dev and 2,352 pairs in test where there are correct answers for questions.

TREC-QA is created by Wang et al. [22] based on Text REtrieval Conference (TREC) QA track (8-13) data. There are two sets of data provided for training. One is the full training set containing 1229 questions that are automatically labeled by matching answer keys' regular expressions. The other one is a small training set contains 94 questions, which are manually corrected for errors. In this paper, the full training set is used because it provides significantly more question and answer sentences for learning, even though some of its labels are noisy. The exact approach of train/dev/test questions selection in [34] is followed, in which all questions with only positive or negative answers are removed.

Answer selection task is to select the only one correct answer corresponding to a question from a candidate answer pool or get them ranked based on their relatedness to the question. The performance is thus measured in Mean Average Precision (MAP) and Mean Reciprocal Rank (MRR), which are standard metrics in Information Retrieval and Question Answering. MRR measures the rank of any correct answer, while MAP examines the ranks of all the correct answers. Therefore, following previous experiments on these three datasets, MAP and MRR are reported in both WikiQA and TREC-QA as evaluation metrics, but accuracy in InsuranceQA because one question only has one right answer in the dataset.

The statistics of these datasets are given in Table 1, including the number of questions, the average number of candidate answers, and the average length of questions and answers.

Table 1. Dataset statistics for InsuranceQA, WikiQA, TrecQA.

| Dataset | InsuranceQA | WikiQA | TrecQA |
|---|---|---|---|
| Train (# questions) | 12887 | 873 | 1162 |
| Dev (# questions) | 1000 | 126 | 65 |
| Test (# questions) | 1800*2 | 243 | 68 |
| Avg (# candidate answers) | 500 | 9 | 38 |
| Avg (#length of questions) | 7 | 6 | 8 |
| Avg (#length of answers) | 95 | 25 | 28 |

*4.2. Baselines*

Four baselines are compared in [7] and [8].
- AP-CNN: An architecture in [7] employs single-layer CNN and basic attention mechanism.
- AP-biLSTM: An architecture combines biLSTM and basic attention mechanism in [7].
- IARNN-Occam: An architecture in [8] pays attention to the original input words and previous hidden state to mitigate attention bias problem.
- IARNN-Gate: An architecture in [8] applies attention deeper to the GRU inner activation.

*4.3. Setup*

**Word Embedding**. The word embedding is trained in the training process. For the InsuranceQA dataset, the 100-dimensional (d = 100) vectors is utilized according to [11]. Following [9], [10] and [34], for the WikiQA and the TREC-QA datasets the 300-dimensional (d = 300) vectors is used. A single randomly initialized embedding is created by uniform sampling from [-0.1,0.1], except [-1,1] for InsuranceQA.

**Hyper-parameters.** All models in this paper are trained and tested using the machine learning library PyTorch. The evaluation metrics (accuracy in InsuranceQA, MAP and MRR in WikiQA and TREC-QA) is utilized on dev set to locate the best hyper-parameter setting. The learning rate in our experiments is 0.001. Adagrad is employed as optimizer. Dropout is further applied to every parameter with probability 30%. The filter size is 3 for single-layer CNN and every layer in multi-layer CNN.

*4.4. Results and Discussion*

In order to explore the performance of MSNN, AM and AM-MSNN, experiments are only conducted with MSNN and AM-MSNN, but also with other three usual neural networks (SingleCNN, MultiCNNs and biLSTM) with combination of AM.

4.4.1. The Effectiveness of MSNN

Here the effectiveness of MSNN is going to be studied by experimental results. The row 2- 4 in Table 2-Table 4 show that MSNN outperforms SingleCNN and MultiCNNs by a large margin for all datasets. This phenomenon can be explained by the impact of different levels of language granularities. For SingleCNN, convolutions capture features only on one level of granularity. For MultiCNNs, note that two-layer CNNs is utilized in experiments because there is not any improvement of two layers of convolution over one layer. Abstract features with larger granularity extracted by the next layer's convolution in MultiCNNs can be influenced directly by the previous layer. Therefore, if the information extracted by the previous layer is not sufficient because of the inappropriate setting of hyper-parameters such as filter size, the number of filters and so on, the effect of using the next layer will be impacted sharply. With the number of layers increasing, this effect becomes more obvious. However, for MSNN, features on different levels of granularities are captured by varied-size filters in parallel, and they will never be impacted by others. In sum, MSNN used in sentence modeling can cover sufficient information from a wider range of granularities than SingleCNNs, and meanwhile, suffer little from their mutual impact, unlike MultiCNNs.

Similarly, MSNN achieves better performance than biLSTM because of the property of answer selection task, comparing row 1 and 4 in Table 2-Table 4. For answer selection task, similar to sentiment analysis, correct answers usually are decided by several words or phrases. Thereby it is difficult to take maximum advantage of long-short distance restriction inside sentences and increase the noise, because biLSTM architecture consists of a cell to record the information before the current step. By the contrast, for MSNN, no matter whether the useful information is included in several word or phrase, it is still a problem about the granularity, which can easily be solved by MSNN.

**Table 2.** Accuracy of different systems without attention for InsuranceQA.

| Idx | System | Dev | Test1 | Test2 |
|---|---|---|---|---|
| 1 | biLSTM | 66.6 | 66.6 | 63.7 |
| 2 | SingleCNN | 61.6 | 60.2 | 56.1 |
| 3 | MultiCNNs | 60.9 | 61.0 | 59.7 |
| 4 | MSNN | 66.9 | 68.0 | 65.1 |

**Table 3.** Performance of different systems without attention for WikiQA.

| Idx | System | MAP | MRR |
|---|---|---|---|
| 1 | biLSTM | 0.6557 | 0.6695 |
| 2 | SingleCNN | 0.6701 | 0.6822 |
| 3 | MultiCNNs | 0.6551 | 0.6721 |
| 4 | MSNN | 0.7164 | 0.7392 |

**Table 4.** Performance of different systems without attention for TREC-QA.

| Idx | System | MAP | MRR |
|---|---|---|---|
| 1 | biLSTM | 0.6750 | 0.7723 |
| 2 | SingleCNN | 0.7147 | 0.8070 |
| 3 | MultiCNNs | 0.7179 | 0.7906 |
| 4 | MSNN | 0.7221 | 0.8051 |

4.4.2. The Superiority of AM

In order to show the superiority of the proposed AM, AM-SingleCNN (single-layer CNN with attention mechanism) and AM-biLSTM (biLSTM with attention mechanism) are compared with AP-CNN (Table 5-Table 7, row 1 vs. 3) and AP- biLSTM (Table 5-Table 7, row 2 vs. 4), respectively. It can be seen that AM achieves better performance by generating more informative representations. Its advantages are that : (1) it reweights outputs of convolution by employing Hadamard production and column-wise max pooling operations and derives the weighted representations which contain the biggest concern for question q with regard to answer a or the one for answer a with regard to question q; (2) it uses concatenation operation over the weighted representations and the outputs of networks to take more information into account.

**Table 5.** Accuracy of different systems with attention for InsuranceQA.

| Idx | System | Dev | Test1 | Test2 |
|---|---|---|---|---|
| 1 | AP-CNN | 68.8 | 69.8 | 66.3 |
| 2 | AP-biLSTM | 68.4 | 71.7 | 66.4 |
| 3 | AM-SingleCNN | 69.6 | 70.5 | 72.7 |
| 4 | AM-biLSTM | 71.3 | 71.5 | 72.4 |

**Table 6.** Performance of different systems with attention for WikiQA.

| Idx | System | MAP | MRR |
|---|---|---|---|
| 1 | AP-CNN | 0.6886 | 0.6957 |
| 2 | AP-biLSTM | 0.6705 | 0.6842 |
| 3 | AM-SingleCNN | 0.7323 | 0.7441 |
| 4 | AM-biLSTM | 0.7195 | 0.7328 |

**Table 7.** Performance of different systems with attention for TREC-QA.

| Idx | System | MAP | MRR |
|---|---|---|---|
| 1 | AP-CNN | 0.7530 | 0.8511 |
| 2 | AP-biLSTM | 0.7132 | 0.8032 |
| 3 | AM-SingleCNN | 0.7531 | 0.8529 |
| 4 | AM-biLSTM | 0.7370 | 0.8301 |

4.4.3. The Superior Performance of AM-MSNN

The superior performance of AM-MSNN is discussed in this section.

Due to the excellent performance of MSNN and AM detailed in 4.4.1 and 4.4.2, AM-MSNN get better performance than AP-CNN, IARNN-Occam, IARNN-Gate and AM-biLSTM.

Attention bias problem is raised in [8], which decreases the effect of attention mechanism. In order to mitigate this problem, IARNN-Occam and IARNN-Gate are proposed and achieve state-of-the-art performance. However, it still is beat by AM-MSNN.

It indicates that multi-size neural network with attention mechanism may be the best architecture applied for answer selection task for the moment.

**Table 8.** Performance of different systems avoiding attention bias problem.

| Idx | System | InsuranceQA | | | WikiQA | | TREC-QA | |
|---|---|---|---|---|---|---|---|---|
| | | Dev | Test1 | Test2 | MAP | MRR | MAP | MRR |
| 1 | IARNN-Occam | 69.1125 | 68.8651 | 65.1396 | 0.7341 | 0.7418 | 0.7272 | 0.8191 |
| 2 | IARNN-Gate | 69.9812 | 70.1128 | 62.7965 | 0.7258 | 0.7394 | 0.7369 | 0.8208 |
| 3 | AP-CNN | 68.8 | 69.8 | 66.3 | 0.6886 | 0.6957 | 0.7530 | 0.8511 |
| 4 | AM-biLSTM | 71.3 | 71.5 | 72.4 | 0.7195 | 0.7328 | 0.7370 | 0.8301 |
| 5 | AM- MSNN | 72.1 | 72.2 | 72.8 | 0.7511 | 0.7569 | 0.7636 | 0.8620 |

*4.5. Visual Analysis for Attention Mechanism*

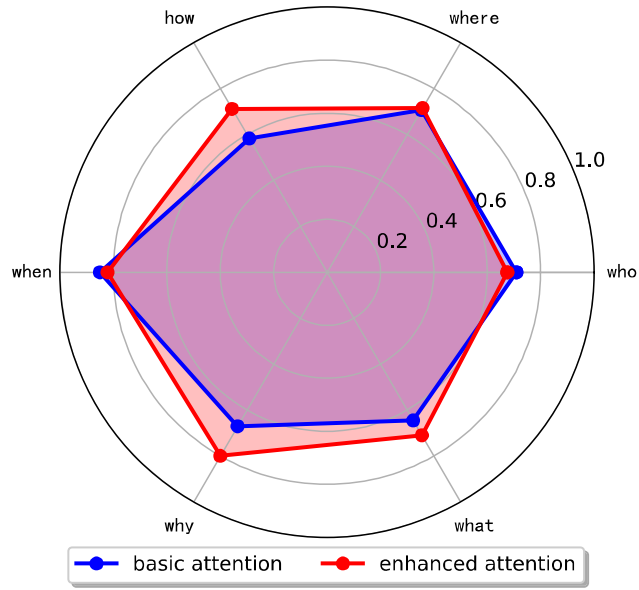

**Figure 4.** Performance of attention mechanism on different types of question.

An experiment is also conducted on WikiQA dataset to measure the effect of AM on different types of questions by comparing with another attention mechanism in [7]. As Figure 4 shows, questions are classified into 6 types (i.e. who, why, how, when, where, what) following the rules mentioned in [8]. As can be seen from the radar graph, AM in this paper achieves similar performance on questions of who, where and when. This indicates that some useful information in questions and answers has been captured by both attention mechanisms. However, AM in this paper beats the one in [7] on types of how, why and what, because AM in this paper generates more informative representations. Therefore, when questions are very clear and definite, such as who, where and when, both perform well. But when questions become fuzzy, representations from feature maps, which contain more information, help AM become more informative, leading to better performance.

*4.6. The Effect of Number of Filters on Performance*

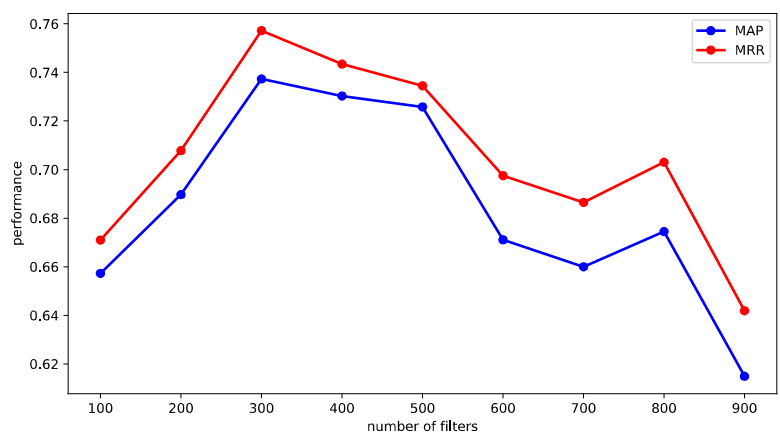

**Figure 5.** Performance according to different numbers of filters size for AM-MSNN on WikiQA

Figure 5 shows the effect of number of filters on performance for the AM-MSNN on WikiQA dataset. As shown in the figure, when the number of filters is 300, it reaches the best performance. One important characteristic of AM-MSNN is that it requires less convolutional filters than AP-CNN (400) in [7]. Because it not only captures feature on different levels of language granularities, but also generates more informative representations.

## 5. Conclusions

In this work, the multi-size neural network (MSNN), an attention mechanism (AM) and their combination (AM-MSNN) are introduced to answer selection task. Experiments are conducted on three different benchmark datasets (InsuranceQA, WikiQA and TrecQA), and experimental results demonstrate that all of them improve the performance by a large margin. And they can be easily adapted to various domains for sentence modeling, such as paraphrase identification (PI) and textual entailment (TE). The main contributions of the paper are: (1) introducing MSNN into answer selection task and analyzing the impact of language granularity for sentence modeling by comparing single-layer CNN, multi-layer CNN and MSNN; (2) proposing AM to generate more informative representations; and (3) combining AM and MSNN to improve the performance of deep learning method for answer selection task. In the future, we plan to explore the performance of AM-MSNN on other tasks.